# A Knowledge Engineer's Comparison of Three Evidence Aggregation Methods

Donald H. Mitchell
Allied-Bendix
9140 Old Annapolis Rd.,
Columbia, MD 21045

Steven A. Harp
Honeywell

David K. Simkin
Northwestern Univ.

The comparisons of uncertainty calculi from the last two Uncertainty Workshops have all used theoretical probabilistic accuracy as the sole metric. While mathematical correctness is important, there are other factors which should be considered when developing reasoning systems. These other factors include, among other things, the error in uncertainty measures obtainable for the problem and the effect of this error on the performance of the resulting system.

There are some domains in which many of the interesting conditional probabilities can be objectively estimated. For example, census data allows various characterizations of individuals with a reasonable degree of confidence. Where information is missing, then a theoretically plausible method, such as maximum entropy assignment, may be used (Cheeseman, 1983).

The typical application domain for artificial intelligence does not have known or accessible objective probabilities. Artificial intelligence systems are being built using knowledge from experts whose knowledge is limited to repeated experience or understanding of major causal factors. Human estimates of probability are known to be biased in several different ways (Edwards, 1971; Fischhoff & Bayth-Marom, 1983; Kahneman & Tversky, 1972; Schum, Du Charme, & DePitts, 1973; Tversky & Kahneman, 1974). The lack of access to accurate objective probabilities leads to the question of how each theory's evidence aggregation technique compound these errors.

Buchanan and Shortliffe (1984) faced this problem of balancing theoretical correctness with domain-specific effectiveness. Because of their experts' difficulties in rationally expressing their knowledge as probabilities, the authors chose an alternative, supposedly more intuitive, evidential metric. Using this metric, they built an effective decision system. Regardless of the success or failure of this and other alternative approaches, it is not known whether using better formulated probabilistic approaches would have performed better or worse. It is possible that the probabilities which the experts were uncomfortable in estimating would have led to better system performance than the comfortable but poorly understood certainty factors.

We conducted an experiment to address this question empirically. In this study, many "expert systems" were created for the same, very simple, problem domain. Each system used a different human expert's judgments and one of three uncertainty representation/aggregation techniques: classical Bayesian, Mycin's certainty factors (Buchanan & Shortliffe, 1984), and Dempster-Shafer (Shafer, 1976) theory. These three were chosen as representative of currently popular methodologies.

The following characteristics were used to design the experiment. First, the three methodologies require that the relevant relations in the domain of expertise be stochastic. Second, the experts had to be comparable in their ability to solve domain problems in order for observed differences to be attributed to differences in the reasoning methodologies. Finally, the actual relations between the evidence and conclusions had to be under experimental control in order to assess the expert systems' accuracies.

Only a highly synthetic domain could satisfy these requirements. We created such a domain and provided extensive controlled experience with it to human experimental subjects. To ensure that the experience would lead to expertise, the domain was kept rather simple; nonetheless, it resembles, in a nontrivial way, the sort of classification task that many expert systems are asked to perform. The subjects predicted the color of blocks based on observed shapes. The experimenter controlled both the conditional probabilities of color given shape and the relative frequencies of different block shapes. This relation is given in Table 1. The relation makes some pieces of evidence (e.g. circles) highly diagnostic while others are worthless (e.g. triangles).

Table 1
Shape Experiment
Number of blocks of each color and shape

|          | Green | Red | Gold | Total |
|----------|-------|-----|------|-------|
| Square   | 0     | 48  | 24   | 72    |
| Circle   | 96    | 16  | 32   | 144   |
| Triangle | 36    | 36  | 36   | 108   |
| Total    | 132   | 100 | 92   | 324   |

The 54 subjects of our study, college students, were required to learn through experience the relation between shape and color. On each of 324 trials, a block was sampled with replacement from the probability distributions. The subjects were shown the block's shape and asked to predict its color. They had to continue guessing until they guessed the correct color. At first their guesses were un-informed; however, as their experience increased, they became more knowledgeable and better able to guess.

The subjects also had to learn how to express their knowledge in terms appropriate to the uncertainty representation used by the system to which they were assigned (Bayesian, Certainty factors, or Dempster-Shafer). Eighteen subjects were assigned to each group. Prior to their exposure to the colored blocks, subjects were given an extensive briefing on how to estimate the appropriate uncertainty parameters. They generated estimates for two domains in which we assumed they had prior experience: the relation between home town and Superbowl viewing, and the relation between economic class and political party affiliation. While learning the colored blocks relation, subjects were asked to make uncertainty parameter estimates at four equally spaced intervals -- i.e., after every 81 trials.

There are many ways to attain the appropriate evidence metrics for each theory. The experiment could not compare each of these methods. Instead, one method was chosen for each of the theories. For the Bayesian metric, the subjects were asked to estimate the a priori probability of each conclu-sion (e.g., what fraction of blocks were red) and the conditional probability of each evidence value given each conclusion (e.g., what fraction of red blocks were square). For the Mycin metric, the subjects were asked to provide a measure of belief change between -1 and +1 for each conclusion given each evidence value (e.g., what change in belief for red knowing the block is square). For the Dempster-Shafer metric, the subjects were asked to estimate a lower and upper bound on the probability of each conclusion given each evidence value (e.g., what fraction of square blocks were red).

Results

The first result of interest is whether the subjects became experts. Each training trial consisted of a shape being shown to the subject and the subject guessing the color until they guessed correctly. At first, the subjects performed no better than chance. On the last 81 trials, however, on those shapes which were diagnostic of color, the subjects performed significantly better than chance, $F(2,34) = 157.04$, $p < .00005$. Table 2 shows the group means and variances. The difference from chance is large enough to consider the subjects expert.

Table 2
Shape Experiment
Average number of guesses per trial on the last set

| shape | Bayes | | Dempster | | Mycin | | Optimal |
|---|---|---|---|---|---|---|---|
| | $\mu$ | $\sigma$ | $\mu$ | $\sigma$ | $\mu$ | $\sigma$ | |
| Square | 1.47 | 0.163 | 1.42 | 0.177 | 1.44 | 0.188 | 1.33 |
| Circle | 1.54 | 0.139 | 1.52 | 0.141 | 1.52 | 0.120 | 1.44 |
| Triangle | 2.06 | 0.152 | 2.00 | 0.178 | 1.97 | 0.159 | 2.00 |

$\mu$ = mean
$\sigma$ = standard deviation

A second result of interest is whether there were any differences in the number of guesses required for subjects assigned to the different evidence aggregation techniques. Subjects in the different groups experienced the same shape-color sequences; only their uncertainty parameter reporting instructions differed. Not surprisingly, no significant difference in guessing ability was observed, $F(2,34) = 2.04$, $p > .14$. This lack of difference allows comparison of the uncertainty parameter estimates because the parameters were obtained from roughly equivalent experts.

Three different methods of comparing knowledge transfer responses were examined. Each method used the performance of expert systems derived from each subject's responses. Briefly, the overall observation from this data is that there were only minor differences in the effectiveness of the different systems. Each comparison will be discussed below.

The first comparison method involved asking each expert system the exhaustive set of questions of the form, "If evidence $i$ is present, is conclusion $x$ more likely than conclusion $y$?" All relevant pair-wise ordinal comparisons were examined. The number of incorrectly ordered pairs measures the inability of each system to discern the conditions which order the likelihoods of the different conclusions. The incorrect responses were totaled.

After the first 81 training trials (the lowest level of learning), the Mycin subjects' responses led to the most accuracy in this task. The Mycin systems answered 85% of these questions correctly while the Dempster-Shafer systems scored 75% and the Bayesian systems scored 70%. This difference was significant and represents possible differences in novices' or intermediate experts' knowledge transfer accuracy.

After all 324 training trials, when the subjects were performing significantly better than chance, the Dempster-Shafer and Mycin systems performed better than the Bayes systems (95%, 93%, and 85% correct respectively), $F(2,34) = 5.75$, $p < .007$.

The evidence indicates that the subjects were better at translating their experience into certainty factors than into Bayesian probabilities. As the subjects became more expert, they became better at estimating the Dempster-Shafer parameters.

The second comparison method tested how well the expert systems derived from the subjects' final phase estimates would do on the same task used to train the subjects. Given a shape, the systems repeatedly guessed the color of blocks until correct. The number of guesses required for each block was tabulated. All systems performed significantly better than chance; however, as with the reversals, the Bayesian systems performed slightly worse than the Mycin or Dempster-Shafer systems, $F(2,34) = 5.57$, $p < .01$.

While the above results support most previous system development designs, they all involve the use of only one piece of evidence to make a decision. The third comparator differs from the previous two in that the expert systems are asked to aggregate multiple pieces of evidence to arrive at a decision. This aspect of performance is certainly important to anyone considering applications of the three uncertainty methodologies. In this scenario, the systems derived from the subjects' final phase estimates diagnose the color of a *bag* of blocks (all the same color) based on a sample of shapes from that bag. Each shape in the sample may be conceived of as an additional piece of evidence. Samples of size 2, 3, 4, 5, 7, 10, 20, 40, and 80 were used.

As the sample size grew, the accuracy of the Bayesian systems increased at a faster rate than the accuracy of either of the other two system groups (see Figure 1). While absolute differences in accuracy grew with sample size, they were nonetheless small – the first to reach statistical significance at the 0.05 level was that between the Bayesian and Mycin groups using a sample size of 40 blocks.

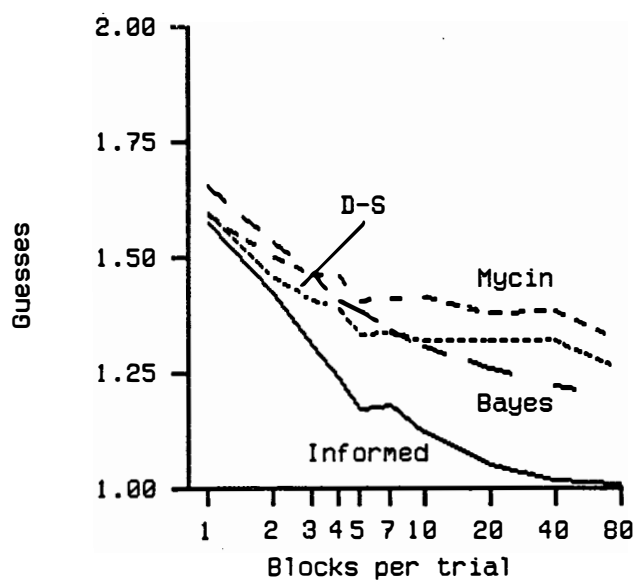

Figure 1: Average number of guesses per trial broken down by expert system group.

As an aside, the Mycin aggregation system appeared to reach asymptotic performance at or about a sample size of five in this task. We cannot offer an explanation for this behavior but observe that it runs contrary to the intuition that more information should lead to improved accuracy.

One interpretation of our experimental results is that psychologically, the correct choice for an uncertainty representation depends on whether the system will be asked to aggregate large collections of evidence or just a few pieces. For large collections, the classical Bayesian approach appears to have an advantage at least over the Mycin system. For small collections, Mycin's certainty factors have an advantage over the Bayesian. The Dempster-Shafer approach appears to do well in both cases.

The results presented here can, by no means, be considered conclusive. The differences, while real, are not vast. In some ways, this lack of a difference is a significant result in the context of Buchanan and Shortliffe's (1984) accepted the conclusion that classical Bayesian parameters cannot be accurately obtained from human experts.

Some factors of this experiment lead to caution in interpreting the results. In particular, we must ask whether in the highly synthetic shape-color problem, people can be expected to acquire and wield judgments of uncertainty that are both qualitatively and quantitatively similar to those acquired over a lifetime of, for example, medical diagnosis. Nonetheless, the differences we have observed lead us to believe that it is worthwhile to be aware of the "human in the loop" and to continue to pursue better ways of acquiring human estimates of uncertainty.

## References

Buchanan, B. & Shortliffe, E. *Rule-Based Expert Systems: The MYCIN experiments of the Stanford Heuristic Programming Project*, Reading, MA: Addison-Wesley, 1984.

Cheeseman, P. A method of computing generalized Bayesian probability values for expert systems. *Proc. Eighth International Conf. on Artificial Intell.*, 1983, 198-202.

Edwards, W. Conservatism in Human Information Processing. *Nature*, 1971, *32*, 414-416.

Fischhoff, B, & Beyth-Marom, R. Hypothesis Evaluation From a Bayesian Perspective. *Psychological Review*, 1983, *90*(3), 239-260.

Kahneman, D. & Tversky, A. Subjective probability: a judgment of representativeness. *Cognitive Psychology*, 1972, *3*, 430-454.

Schum, D. A., Du Charme, W. M., & DePitts, K. E. Research on Human Multistage Probabilistic Inference Processes. *Organizational Behavior and Human Performance*, 1973, *10*(3), 318-348.

Shafer, G. *A Mathematical Theory of Evidence*. Princeton University Press, 1976.

Tversky, A. & Kahneman, D. Judgment under uncertainty: Heuristics and biases. *Science*, 1974, *185*, 1124-1131